\documentclass[letterpaper, 10 pt, conference]{ieeeconf}  

\IEEEoverridecommandlockouts                              

\usepackage{graphics} 
\usepackage{amsmath} 
\usepackage{graphicx}
\usepackage{svg}
\usepackage{url}
\usepackage{xcolor}
\usepackage[linesnumbered,ruled,vlined]{algorithm2e}

\renewcommand{\baselinestretch}{0.96} 

\title{\LARGE \bf
FRIDA: A Collaborative Robot Painter with a Differentiable, Real2Sim2Real Planning Environment
}

\author{
    Peter Schaldenbrand$^{1}$, James McCann$^{1}$, and Jean Oh$^{1}$%
    \thanks{
        $^{1}$The Robotics Institute, Carnegie Mellon University
    }
    \thanks{
        \{pschalde, jmccann, hyaejino\}@andrew.cmu.edu
    }
}\vspace{-7pt}

\begin{document}

\maketitle
\thispagestyle{empty}
\pagestyle{empty}

\vspace{-10pt}
\begin{abstract}
    Painting is an artistic process of rendering visual content that achieves the high-level communication goals of an artist that may change dynamically throughout the creative process.
    In this paper, we present a Framework and Robotics Initiative for Developing Arts (FRIDA) that enables humans to produce paintings on canvases by collaborating with a painter robot using simple inputs such as language descriptions or images. FRIDA introduces several technical innovations for computationally modeling a creative painting process. 
    First, we develop a fully differentiable simulation environment for painting, adopting the idea of real to simulation to real (real2sim2real). We show that our proposed simulated painting environment is higher fidelity to reality than existing simulation environments used for robot painting. 
    Second, to model the evolving dynamics of a creative process, we develop a planning approach that can continuously optimize the painting plan based on the evolving canvas with respect to the high-level goals.  
    In contrast to existing approaches where the content generation process and action planning are performed independently and sequentially, FRIDA adapts to the stochastic nature of using paint and a brush by continually re-planning and re-assessing its semantic goals based on its visual perception of the painting progress.
    We describe the details on the technical approach as well as the system integration. FRIDA software is freely available at: \url{https://pschaldenbrand.github.io/frida/}. 
\end{abstract}

\section{INTRODUCTION}

Painting is the artistic process of rendering visual content that achieves an artist's high-level, semantic goals. As opposed to its straightforward analogue printing--i.e., creating a copy of an original input--painting is a dynamic process where an artist's initial goals are generally vaguely defined and may change dynamically during the creative process~\cite{hertzmann2022algorithmicPainting,walia2019dynamicCreativity} and the artist's goals are specified semantically and at a high-level.

\begin{figure}[t]
    \centering
    \small
     \includegraphics[width=\columnwidth]{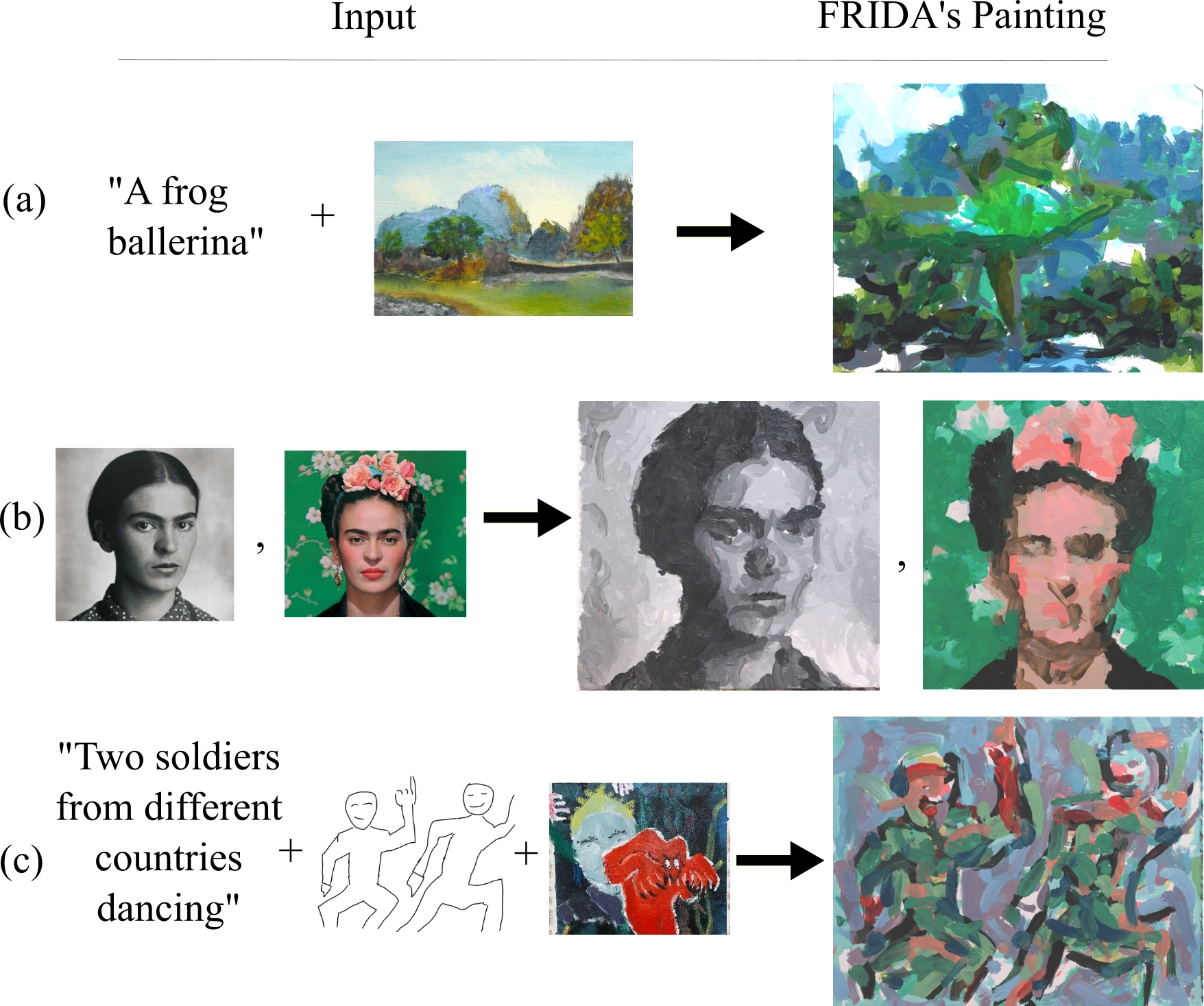}\vspace{-8pt}
     \caption{Human users interacting with FRIDA can describe semantic goals using a variety of simple input such as (a) natural language and style images, (b) source images to be painted precisely, and (c) sketches with language description and style images. Input image sources: \cite{landscapePainting, fridaBW, fridaColor, troyImage}.
     }\label{fig:example_results}
\end{figure}
\begin{figure}[!h]
    \centering
    \small
    \includegraphics[width=6.5cm]{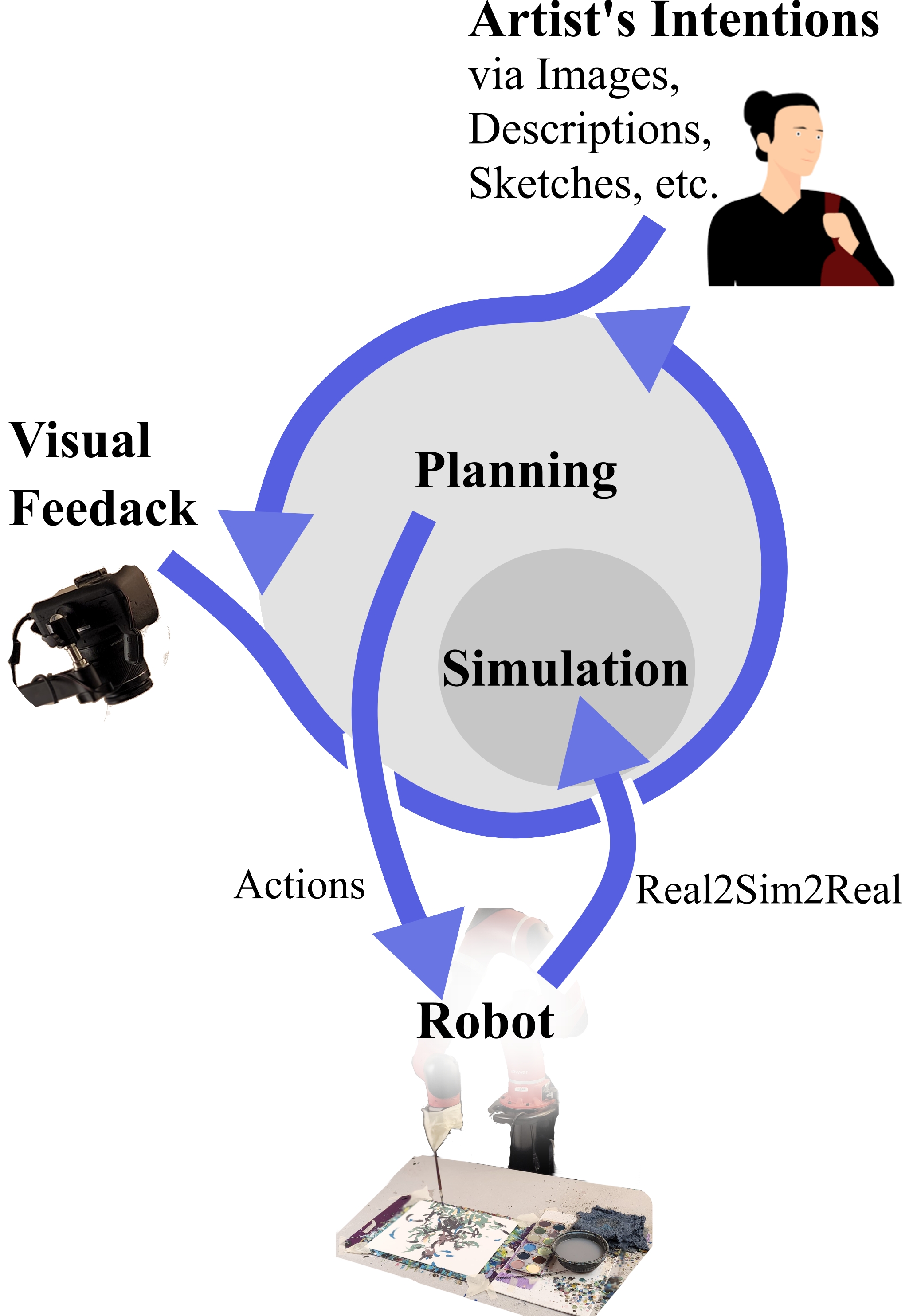}
    \caption{Planning, in FRIDA, is designed to achieve the high-level goals of the human user, which are specified with multi-modalities. FRIDA plans in a simulated environment created from real robot data and continually optimizes its plan based on visual perception to ensure the artist's intentions are realized.}
    \label{fig:workflow}\vspace{-20pt}
\end{figure}
\begin{figure}[!h]
    \centering
    \small
    \includegraphics[width=7cm]{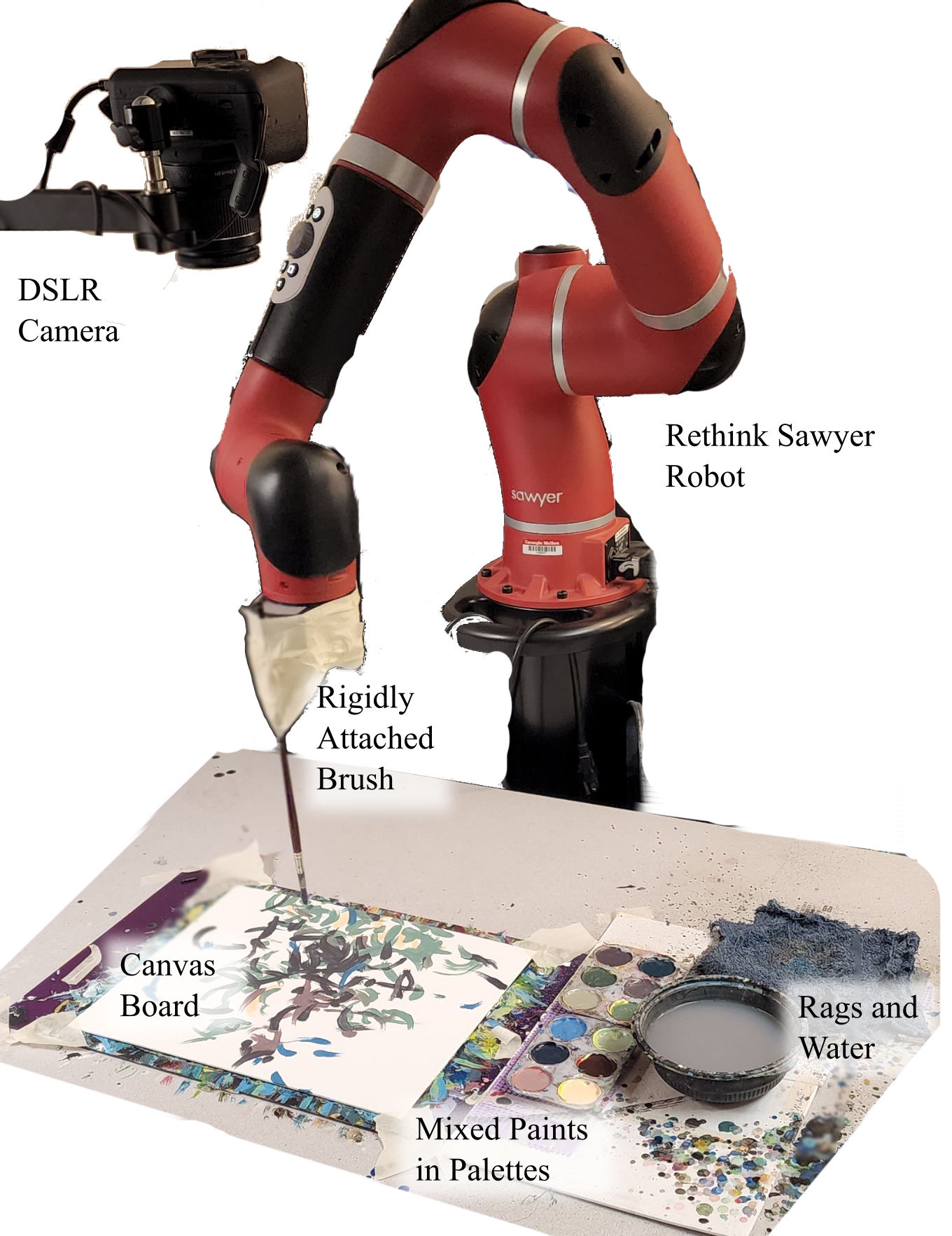}\vspace{-8pt}
    \caption{FRIDA's embodiment and workspace.}
    \label{fig:setup}\vspace{-20pt}
\end{figure}

In this paper, we consider how we can enable robots with painting abilities.
To this end, we introduce a Framework and Robotics Initiative for Developing Arts (FRIDA) that can computationally model creative processes such as painting to support human creativity. FRIDA showcases a fully integrated robotic system (Fig. \ref{fig:workflow}) that can take inspirations from a human user in the text or image format, e.g., ``two soldiers from different countries dancing,'' to produce an artistic painting (Fig.~\ref{fig:example_results}).
FRIDA does so by modeling painting as a \textit{planning} problem where a canvas constitutes the state space and brush strokes are available actions. Here, the objective is to find a sequence of actions, e.g., brush strokes, that would result in a final canvas that satisfies an artist's goals or intentions.  An additional objective is to allow for the adjustment of goals based on the progressive execution of a plan. Contrary to existing robot painting systems where a fully programmed sequence of actions is blindly executed by a robot, FRIDA's painting approach interleaves content generation and action planning to achieve continual content optimization.


FRIDA's core technologies draw on two insights of the artistic process~\cite{hertzmann2022algorithmicPainting}: 1) art has high-level, semantic goals, and 2) art is a dynamic process which needs to adapt and reconsider its goals constantly during the creation process.  
To achieve high-level semantic goals, we design loss functions to compare the semantic goals (or user inputs) and the current canvas. Instead of computing the loss using the pixel values, we use the feature values from pre-trained deep neural networks (DNNs) as the features extracted from DNNs tend to correlate well with human perception of image style~\cite{kolkin2019-strotss, gatys2016image, schaldenbrand2022styleclipdraw} and multi-modal alignment~\cite{frans2021-clipdraw, smith2021-clipGuided}.
We hypothesize that using DNN features would result in paintings that are semantically relevant to high-level goals as opposed to merely replicating (or printing) a given input.  
In order to plan using feedback from DNNs, we create a differentiable, simulated painting environment which enables our planner to optimize brush actions directly toward semantic goals using stochastic gradient descent.  Inspired by the idea referred to as Real2Sim2Real used in other robotics problems~\cite{lim2022real2sim2real}, the simulation environment is created by modeling real brush strokes generated by the robot which reduces the Sim2Real gap.

Capturing the dynamic nature of painting is a challenge when framing it as a robot painting problem.
The majority of existing work in robot painting models the painting process as analogue printing, that is, an input image is the same as their final goal to reproduce in painting. Departing from those views, our approach follows the idea of planning with partial information in robotics where an initial plan is generated to initiate the execution but continuously gets updated as a robot acquires more information from an environment. 

Our contributions: 
\begin{enumerate}
    \item A robotics framework and initiative for interdisciplinary research to promote human creativity, addressing technical challenges in core robotics fields including  planning, simulation,
    and human-robot interaction in the domain of visual arts. 
    FRIDA's complete software stack is open sourced  (\url{https://github.com/pschaldenbrand/Frida}); 
    \item A differentiable, high-fidelity painting simulation environment developed using Real2Sim2Real methodology that reduces the Sim2Real gap from prior works;
    \item A planning algorithm for performing multiple tasks that have high-level, semantic goals under stochastic circumstances;
    \item An intuitive interaction interface, e.g., human users can describe semantic goals using natural language, sketches, and/or style images.
\end{enumerate}\vspace{-5pt}

    \begin{figure*}[h]
        \centering
        \includegraphics[width=\textwidth]{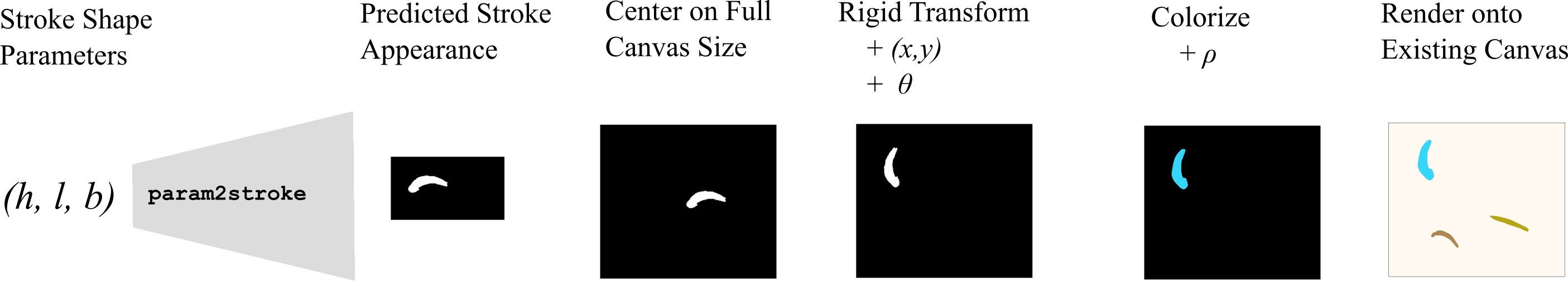}
        \caption{The process of rendering a stroke, given its parameters, onto an existing canvas in our differentiable simulated painting environment.}
        \label{fig:stroke_rendering_process}\vspace{-15pt}
    \end{figure*}
    
    \begin{figure}
        \centering
        \includegraphics[width=\columnwidth]{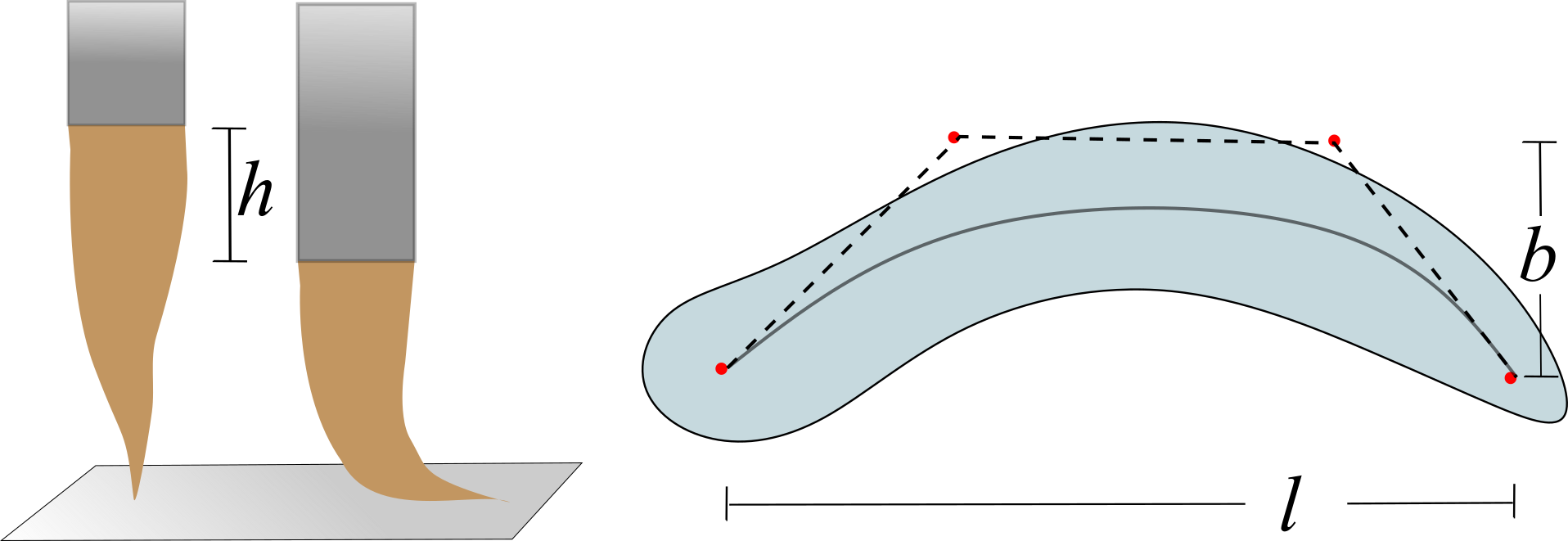}\vspace{-8pt}
        \caption{Our brush shape model has three parameters: thickness ($h$), bend ($b$), and length ($l$).}
        \label{fig:brush_shape_model}\vspace{-15pt}
    \end{figure}

\section{RELATED WORK}

\subsection{Simulated Painting}\label{sec:related-sim} 
    Stroke-Based Rendering (SBR) recreates a given target image using a set of primitive elements that usually resemble brush strokes of paint. 
    Procedural SBR methods generally use rules and heuristics to generate the stroke plan \cite{hertzmann1998painterly, zeng2009imageParsingPainterly}. Planning-based SBR methods use search, optimization, or learning models such as Reinforcement Learning or Recurrent Neural Networks to generate a stroke plan with an objective of replicating an input image~\cite{hertzmann2001paintByRelaxation, huang2019learningToPaint}. 
    
    Recent SBR methods expands the input space to incorporate high-level goals to generate brush stroke simulated paintings based on language descriptions and/or style specification~\cite{frans2021-clipdraw, schaldenbrand2022styleclipdraw, zou2021stylizedNeuralPainting}.
    While these methods present appreciable results in simulation, 
    technical challenges specific to transitioning from simulation to real robots have not been addressed.
    
\subsection{Robot Painting}
There have been numerous robot-created paintings including notable works that had competed in an annual competition in 2016--2018 \cite{robotArtCompetition}, but technical details of most works have not been published. Based on published works, existing robot painting approaches can roughly be categorized into two groups: engineered systems and learning-enabled systems.

    \subsubsection{Engineered robotic painting systems} use well measured equipment to ensure that the planning environment is accurate to the real environment and use rules and heuristics for planning. The Dark Factory portraits~\cite{carter2017DarkFactoryPortraits} utilize a highly accurate robotic arm with known models of brush shape and size. They plan a full sequence of actions a priori such that the plan can be blindly executed.  E-David~\cite{lindemeier2018Edavid} uses a simulated environment constructed to be similar to its painting equipment then draws strokes perpendicular to gradients in the target image.
    In general, the engineered systems are capable of high-fidelity reproductions of input images as they meticulously engineer to minimize the sim2real gap in their setup; however, they are not generalizable to different equipment or settings. Furthermore, these approaches generally do not support more than replicating a given image. 
    
    \subsubsection{Learning-enabled robotic painting systems} generally use simulation environments to plan brush strokes and then execute the plan using a physical robot. 
    Due to a huge sim2real gap, brush stroke plans based directly on simulation methods~\cite{frans2021-clipdraw,schaldenbrand2022styleclipdraw} produce poor-quality paintings or are even infeasible on real robot systems. It has been shown that additional constraints help reducing the sim2real gap to enable robots to paint according to a generated plan~\cite{schaldenbrand2021contentMaskedLoss,sola2022dreamPainter}, but such rigid constraints sometimes result in vague or imprecise outcomes. 
    In line-drawing, \cite{lee2022icraSketchRobot} used reinforcement learning to learn both the SBR instructions and the low level robot instructions for the reproduction of sketches. Their approach is designed to plan once and execute a given plan as is without observation feedback in the loop. In painting, however, visual feedback is crucial as painting is a continuously evolving process~\cite{hertzmann2022algorithmicPainting}.



\subsection{Brush Stroke Modeling}

Brush strokes can be represented using a height map and a color map as in~\cite{wu2018deepBrush} where Generative Adversarial Networks are used to map a user input trajectory into a synthesized brush stroke. In their work, both training and testing were done using data synthesized using a volumetric oil painting simulator based on WetBrush~\cite{chen2015wetbrush}. While the outputs appear impressive in simulation, the challenge still remains unanswered how such a simulated input can be translated into a real painting, for example, by a robot.

Wang et al.~\cite{wang2020calligraphy} use brush parameters such as the width, drag, and offset of the brush's bristles to create a very accurate brush stroke model. 
They use pseudospectral optimal control to optimize trajectories of brush strokes to fit the target calligraphy character, which works well with calligraphy where an initial path is given in a reasonably accurate form and the brush strokes are clearly separated by white space. In the painting domain, however, a more generalizable approach is needed due to the fact that the shapes of brush strokes used in painting are highly flexible and unconstrained and that brush strokes frequently overlap with previous ones. 


\section{METHODS}

    \subsection{Brush Stroke Model}\label{sec:brush_shape_model}
    
    Inspired by \cite{wang2020calligraphy}, we parameterize the space of brush strokes using three parameters as shown in Fig.~\ref{fig:brush_shape_model}. 
    In addition to brush shape attributes, i.e., 
    the length $l$ of the stroke, and the amount $b$ that the stroke bends up or down, 
    the thickness $h$ of the stroke specifies how far the brush is pressed proportionally to the canvas. 
    A brush stroke is parameterized by its shape, denoted by ($h, l, b$), the location coordinates on a canvas ($x,y$), orientation $\theta$, and color $\rho$ in the RGB format. The stroke trajectory can then be represented by a cubic B\'ezier curve where the horizontal coordinates are a linear interpolation between $0$ and $l$, and the vertical coordinates are $0$ at the end points and $b$ in the center points.

    \subsection{Real Data to Simulation}
    Whereas existing models such as~\cite{wang2020calligraphy} use only shape features of the rendered images that would require some model of a brush tool for a robot control interface, our definition of thickness connects the parameter space with a brush tool and a robot. During the calibration phase, we generate random brush strokes to train the \texttt{param2stroke} model, a Neural Network comprised of two linear layers followed by two convolutional layers and an bilinear upscaler, that translates a brush stroke shape tuple directly into the appearance map of the brush stroke.  The brush stroke shape tuple can deterministically be translated into control inputs for a real robot.
    
    A rudimentary approach to creating a differentiable, simulated robot painting environment would be to allow the robot to paint randomly and continuously to collect a large enough dataset of paired robot actions to the effects on the canvas to model this relationship. While this method works well in simulated environments~\cite{wu2018deepBrush, schaldenbrand2021contentMaskedLoss, huang2019learningToPaint, zou2021stylizedNeuralPainting}, where thousands of brush strokes can be produced on the order of seconds, generating a similarly large-sized real dataset is impractical. Painting in real life is slow, and if the brush or other materials were altered, the entire process would need to be restarted. 
    Instead, we augment the dataset using existing differentiable functions, such as rigid transformations for positioning and orienting strokes and stamping methodology for rendering individual brush strokes onto an existing canvas, to allow our painting environment to be simulated with a reasonably small number of real brush strokes for modeling.
    
    \subsection{Differentiable Simulated Painting Environment}
    
    The stroke rendering process in our simulation is depicted in Fig.~\ref{fig:stroke_rendering_process}: the \texttt{param2stroke} network translates the thickness, bend, and length parameters into a 2d magnitude map of the brush stroke's predicted appearance. This magnitude map is then padded such that it is the size of a full canvas.  Then the map is translated and rotated to the specified orientation and location. The magnitude map is converted into an RGBA image, and then the stroke is applied to a given existing canvas.  Strokes can be layered upon each other to create a complete simulated painting. They can also be rendered onto a photograph of the real canvas for planning throughout the painting process. The whole rendering process is differentiable, meaning that the loss value computed using the rendered canvas can be differentiated, back-propagated through the simulator, and a Stochastic Gradient Descent algorithm updates the brush stroke parameters such that the parameters minimize the loss function.
    
    \subsection{Planning Algorithm} \label{sec:planning_algorithm}
    
    Algorithm \ref{alg:painting_algorithm} sketches our planning algorithm for painting with visual feedback given high-level semantic goals. 
    The \texttt{optimize} function compares the goals (image or text) with the simulated painting according to one or more objective functions detailed in Section \ref{sec:objective_functions}.  The simulated painting is the rendering of the brush stroke plan onto the current canvas--i.e., the current image observed by the camera sensor. 
    
    To start the \texttt{paint} function, $n\_strokes$ brush strokes are randomly initialized by sampling uniformly over the brush stroke parameters. An initial pass at optimization can be performed with initial high-level goal(s), then the plan is optimized to use the full objective(s). The first $batch\_size$ brush strokes from the plan are executed, then the remainder of the plan is re-optimized based on the executed brush strokes and objective(s).  
    

    \begin{algorithm}[t]
    \caption{Painting Planning Algorithm}
    \label{alg:painting_algorithm}

    \SetKwProg{Fn}{Def}{:}{}
    \SetKwFunction{FMain}{Main}
    \SetKwFunction{FSum}{Sum}
    \SetKwFunction{FSub}{Sub}
    \SetKwFunction{FPaint}{paint}
    \SetKwFunction{FOptimize}{optimize}
    \Fn{\FOptimize{$plan$, $targets$, objectives}}{
        canvas = camera() \\
        \While{plan \textrm{is not} optimized}
        {
            sim\_painting = canvas + plan.render()\\
            loss = 0\\
            \For{objective, target \textrm{in} objectives, targets}
            {
                loss += objective(sim\_painting, target)
            }
            \textrm{\small \# Update plan to decrease loss with SGD}\\
            plan.update(loss) 
        }
        \KwRet plan
    }
    \Fn{\FPaint{$targets$, $n\_strokes$, objectives, init\_objectives}}{
        \textrm{\small \# Initialize a brush stroke plan}\\
        $plan$ = \texttt{init\_brush\_strokes}($n\_strokes$)\\
        \textrm{\small \# [optional] Do an initial pass at optimization}\\
        \If {init\_objectives \textrm{is not} None} {
            plan = optimize(plan, targets, init\_objectives)
        }
        plan = optimize(plan, targets, objectives)\\
        \While{plan.n\_strokes $>$ 0}
        {
            \textrm{\small \# Paint some strokes from plan}\\
            plan.execute(batch\_size) \\
            plan = plan[batch\_size:] \\
            \textrm{\small \# Update Plan}\\
            plan = optimize(plan, target, objective)
        }
    }
    \end{algorithm}

    \subsection{High-Level Goal Objective Functions}\label{sec:objective_functions}
   
    To enable our system to achieve high-level goals, we employ a variety of objective functions from recent image synthesis literature. Each objective function has a loss function that compares the brush stroke plan ($p$) to the target input ($t$) which may be language or an image.  A plan $p_{next}$ for the next time step is rendered into a raster image using a differentiable simulated environment ($r$). These objective functions can be used in different combinations to achieve high-level, artistic tasks, e.g., 
        painting from language description with or without a specified style, 
        painting images conceptually,
        or
        painting from a sketch.
        
        \begin{figure}[!t]\vspace{-12pt}
        \begin{align}
            &l_{1} = l_{text} =  \cos(CLIP_{img}(r(p)), CLIP_{text}(t)) \label{eq:img_text}\\
            &l_{2} = l_{style} = EMD(VGG(r(p)) - VGG(t)) \label{eq:style} \\
            &l_{3} = l_{print} =  ||r(p) - t||^{2}_{2} \label{eq:l2} \\            
            &l_{4} = l_{semantic} =  ||CLIP_{conv}(r(p)) - CLIP_{conv}(t)||^{2}_{2}\\
            &p_{next} = \min_{p} \sum_{i=1}^{4}{(w_{i} l_{i})}, w_{i} \in \{0, 1\}
            \label{eq:clip_conv_loss}
        \end{align}\vspace{-25pt}
        \end{figure}
    
        \noindent\textbf{Image-Text Similarity Objective} (Eq. \ref{eq:img_text})
        This objective optimizes the brush stroke plan ($p$) such that the cosine distance between the CLIP~\cite{radford2021-clip} embeddings of both the painting and the language description ($t$) is minimized, guiding the painting to resemble the content of the text, as is common in recent CLIP-guided text-to-image synthesis methods~\cite{schaldenbrand2022styleclipdraw, frans2021-clipdraw, Galatolo2021-ClipStyleGAn2, smith2021-clipGuided}.
    
        \noindent\textbf{Style Objective} (Eq. \ref{eq:style})
        Given an example style image, the style objective guides the painting to resemble the colors, shapes, textures, and other style features of the given image. This objective was created for style transfer methodology~\cite{kolkin2019-strotss, gatys2016image}. The style objective minimizes the Earth Mover's Distance ($EMD$) between style features that are extracted using a pretrained object detection model ($VGG$~\cite{simonyan2014-vgg16}), from the brush stroke plan ($p$) and the style image ($t$).
        
        \noindent\textbf{Simple Replication Objective}  (Eq. \ref{eq:l2})
        Image replication is not considered a high-level goal. Instead, it is a straightforward minimization of the $L_2$ distance between the rendered brush stroke plan and the target image ($t$).

        \noindent\textbf{Semantic Replication Objective}  (Eq. \ref{eq:clip_conv_loss})
        Following \cite{vinker2022clipasso}, features can be extracted from the convolutional layers of CLIP which are rich in both semantic and geometric information. 
        For a high-level semantic replication objective, we minimize the $L_2$ difference of features extracted from the target image and painting from the last convolutional layer of CLIP ($CLIP_{conv})$.

\section{Robot Setup Details}

    We used a Rethink Sawyer robot~\cite{rethinkSawyer} as a machine to test our approach. Any Robotics Operating System (ROS) compatible machine with a similar morphology to the Sawyer could feasibly be adapted to execute our approach with only minimal changes to the robot interface code.
    
    A photograph of our painting equipment and setup can be seen in Fig. \ref{fig:setup}. We use a Canon EOS Rebel T7 to perceive the canvas. For all examples in this paper, we used $11\times14$ inch canvas board as painting surfaces.  Premixed acrylic paints are provided to the robot in palette trays with up to 12 color options available. 
    Alternatively, from an initial painting in simulation, the colors are discretized to a user-specified number using K-Means cluster; palette preparation is performed accordingly by a human.
    A rag and water are provided for the robot to clean paint off of the brush, which is performed when switching colors. The brush is rigidly attached to the robot's end effector and is always held perpendicular to the canvas.  Indirect, diffused lighting is necessary, since direct lighting can cause too much glare from the wet paint into the camera.  
    The locations of all the painting materials (canvas, paint, water rag) with respect to the robot are explicitly programmed.


\section{RESULTS}


\begin{figure}
    \centering
    \small
    \includegraphics[width=\columnwidth]{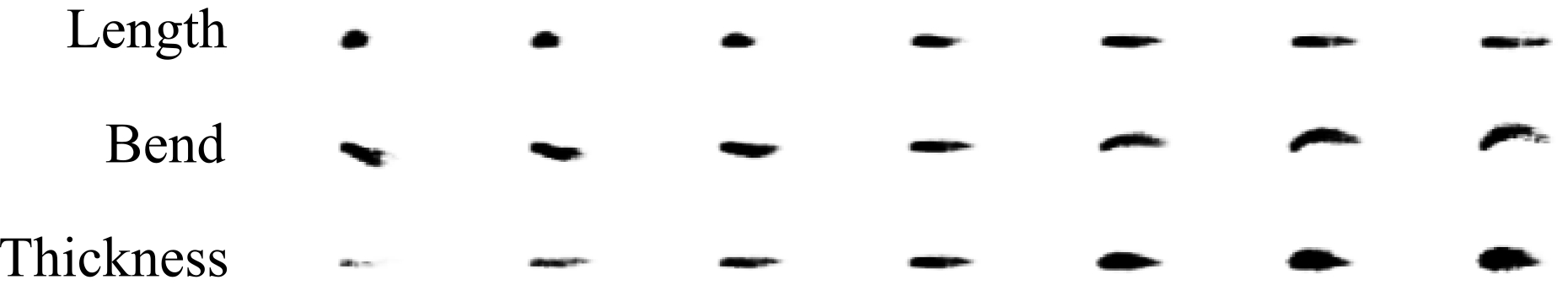}
    \caption{Depictions of interpolating between minimum and maximum values of each of the three stroke shape parameters with the trained \texttt{param2stroke} model.}
    \label{fig:continuous_strokes}\vspace{-7pt}
\end{figure}

\begin{figure}
    \centering
    \includegraphics[width=\columnwidth]{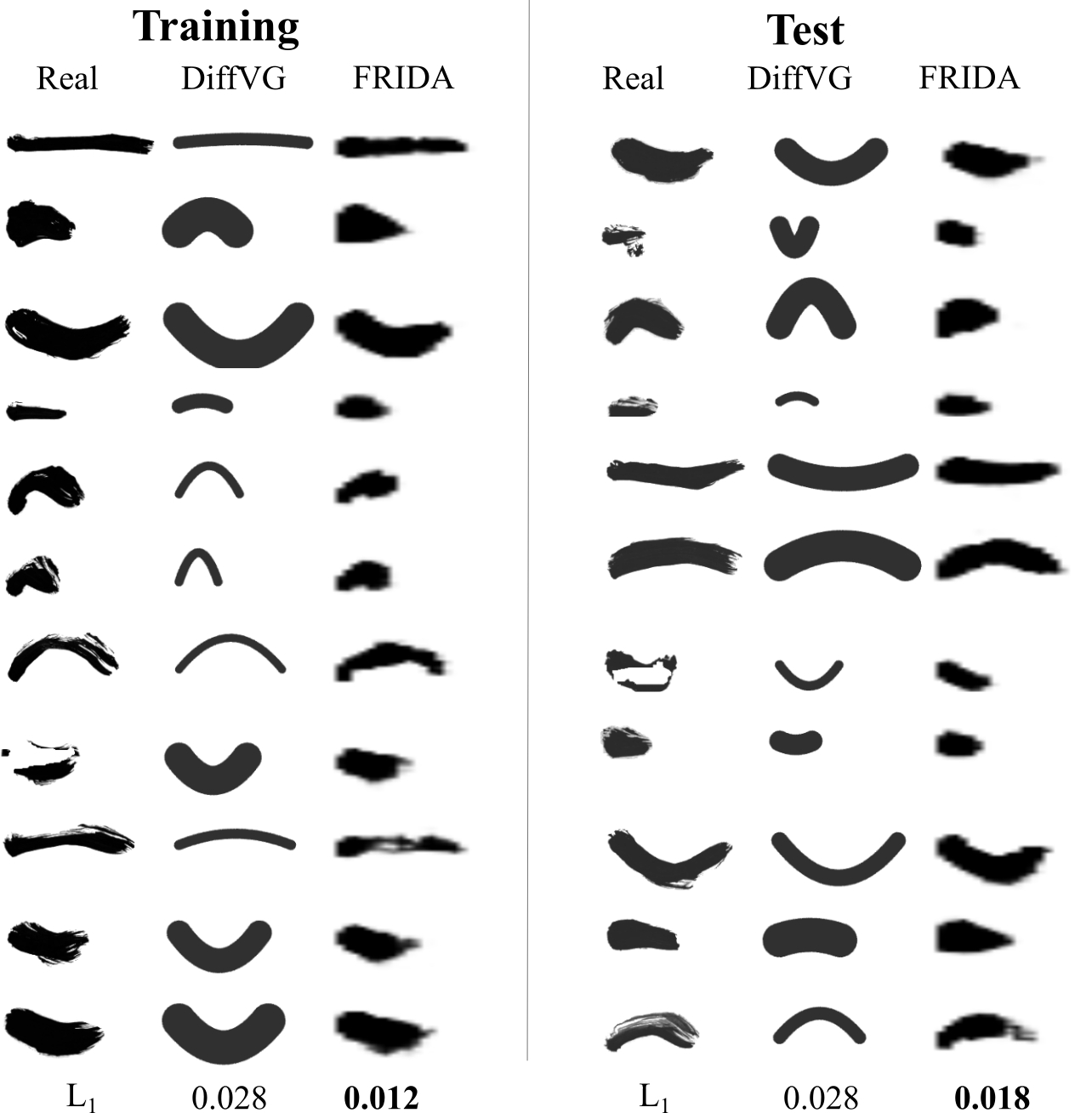}
    \caption{We compare using DiffVG~\cite{li2020-diffVG} and FRIDA's \texttt{param2stroke} model for modeling brush stroke shapes. The average $L_1$ distance computed on 50 samples between the modeled and real brush strokes is displayed at the bottom.}
    \label{fig:real_v_sim_strokes}\vspace{-15pt}
\end{figure}

\subsection{Simulated Painting Environment}

The trained \texttt{param2stroke} model can produce strokes with continuous parameter values as seen in Fig.\ref{fig:continuous_strokes}.
Fig.~\ref{fig:real_v_sim_strokes} shows the difference between real brush strokes and FRIDA's modeled brush strokes using the same input parameters. We also compare these strokes to DiffVG~\cite{li2020-diffVG} which was used for brush stroke planning in~\cite{sola2022dreamPainter}. The average $L_1$ loss between the modeled and real strokes was significantly (p-value $< .01$) less for FRIDA's stroke model than DiffVG.

\begin{figure}
    \centering
    \small
    \includegraphics[width=\columnwidth]{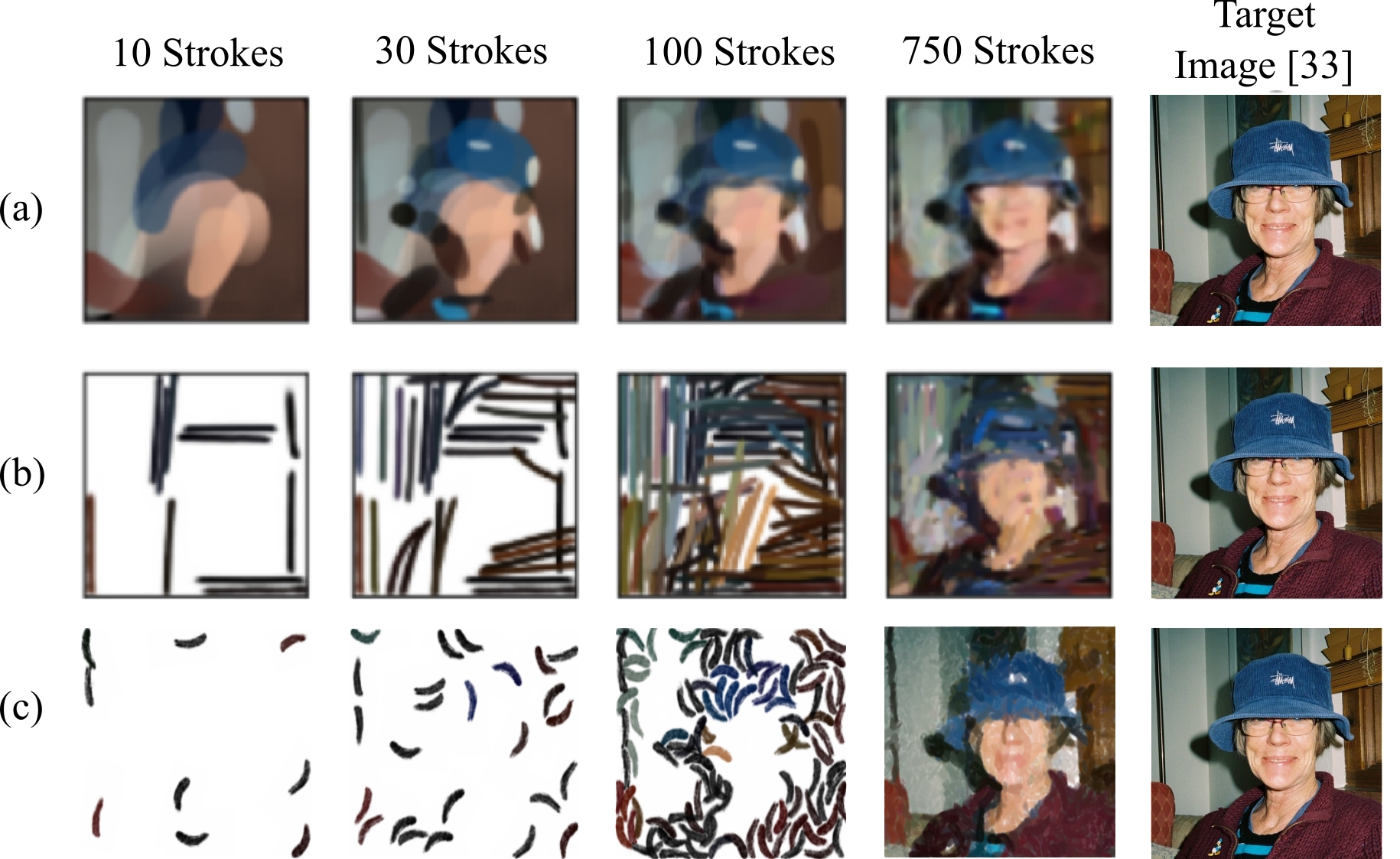}
    \caption{Comparing the simulation environments of three painting methods painting with various numbers of brush strokes: (a) Huang et al. 2019 \cite{huang2019learningToPaint}, (b) Schaldenbrand \& Oh 2021 \cite{schaldenbrand2021contentMaskedLoss}, (c) FRIDA (proposed)}
    \label{fig:vs_baseline_sim}\vspace{-7pt}
\end{figure}

We qualitatively compared our approach to 
Huang et al. 2019~\cite{huang2019learningToPaint} and Schaldenbrand \& Oh 2021~\cite{schaldenbrand2021contentMaskedLoss}.
Fig.~\ref{fig:vs_baseline_sim} compares the brush strokes in early stages of painting simulation where we can observe drastic differences.

\begin{figure}
    \centering
    \small
    \includegraphics[width=\columnwidth]{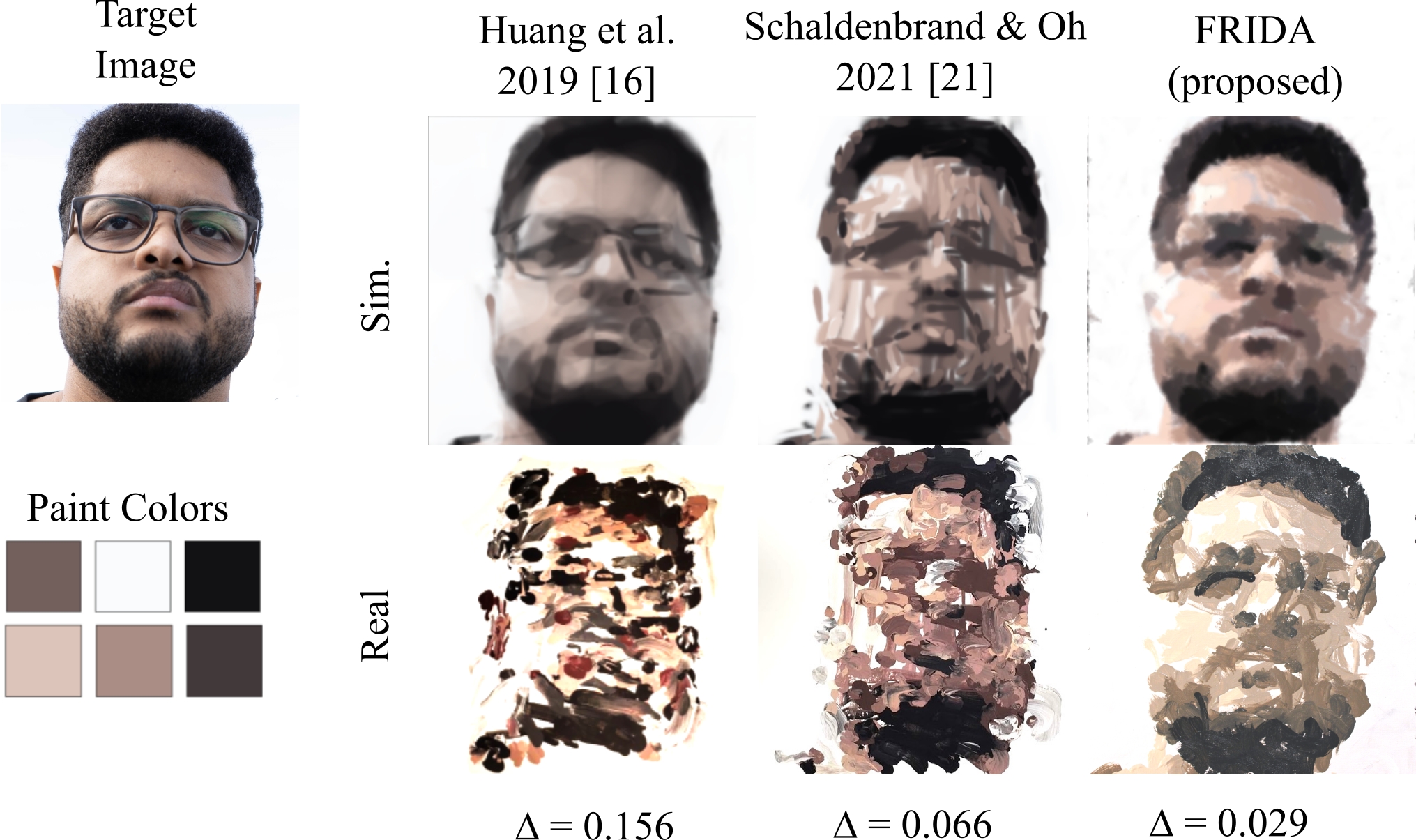}\vspace{-8pt}
    \caption{We compare the Sim2Real gap between FRIDA and two existing methods. The MSE between the simulated plan and the real painting is displayed below each pair.}
    \label{fig:vs_baseline}\vspace{-15pt}
\end{figure}

Fig.~\ref{fig:vs_baseline} shows the comparison in terms of the sim2real gap for entire paintings. 
In simulation, Huang et al. 2019's Reinforcement Learning (RL) model is able to almost perfectly replicate a given image due to their unconstrained stroke model, e.g., allowing strokes that are huge in size and have varying opacity; however, when we fed the strokes to a painting robot, the painting produced was vastly dissimilar to both the simulation and target image.
Schaldenbrand \& Oh 2021 constrained the brush stroke parameters (length, width, color, and opacity) such that a robot was more capable of executing the strokes; however, the constraints made it challenging for their RL model to accurately replicate a target image.  For the proposed approach, we used our simulation to recreate the target image using the Simple Replication Objective (Eq. \ref{eq:l2}) and did not re-plan with perception for fair comparison. Our proposed approach shows clearly visible improvement in recreation both in simulation and real painting.

    \begin{figure}[t]
        \centering
        \includegraphics[width=\columnwidth]{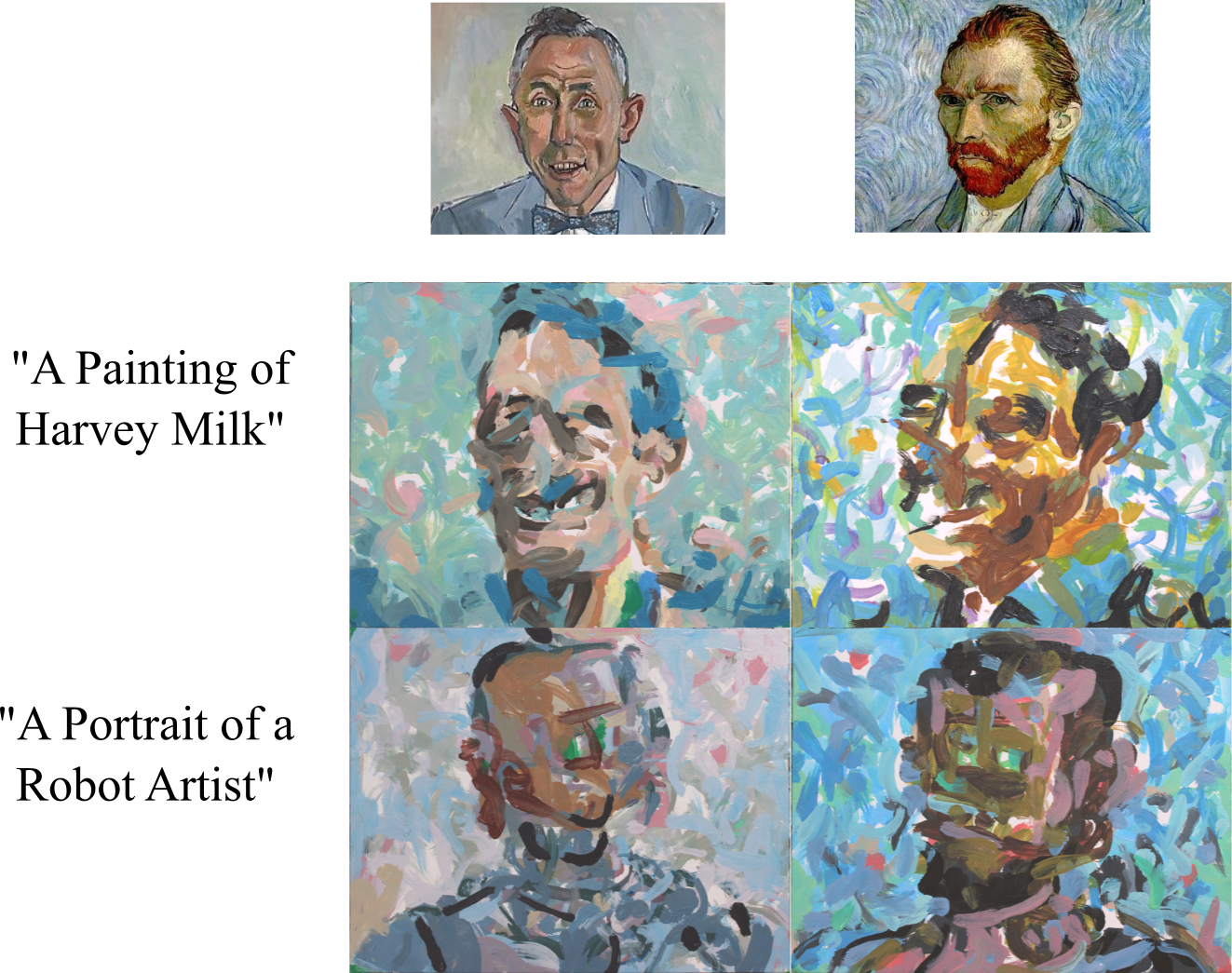}
        \caption{
        We replicate StyleCLIPDraw~\cite{schaldenbrand2022styleclipdraw}, a method for generating drawings using language descriptions and style from example images, using FRIDA.
        }
        \label{fig:style_and_text}\vspace{-7pt}
    \end{figure}

\subsection{Dynamic Planning and Adaptation}
    
\begin{figure}
        \centering
        \small
        \includegraphics[width=\columnwidth]{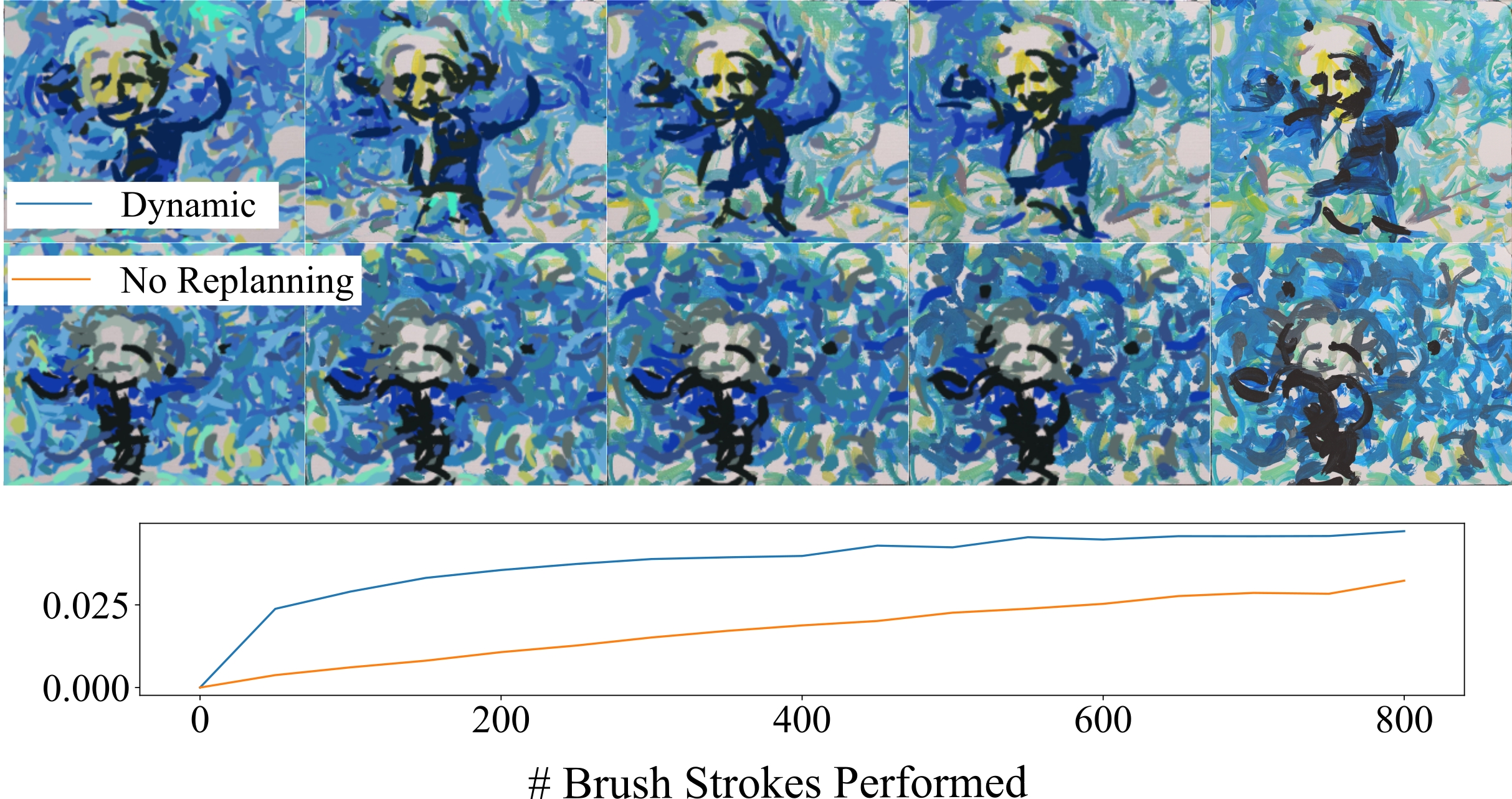}\vspace{-8pt}
        \caption{FRIDA painting with text input ``Albert Einstein Dancing" in the style of van Gogh's \textit{The Starry Night} with and without replanning. The left most images are the initial plan followed by the plan after 200 brush strokes performed. Below, the mean squared error between the current plan versus the initial plan is plotted.}
        \label{fig:init_plan_deviation}\vspace{-20pt}
    \end{figure}

    We painted with and without FRIDA's dynamic replanning system and plotted the deviation from the initial plan in Fig. \ref{fig:init_plan_deviation}. Without replanning, the difference between the current and initial plan grows linearly as the plan is executed from simulation to reality stroke by stroke.  With replanning, the plan changes more significantly from the initial plan as the algorithm adapts to the stochastic  execution of the plan, resembling the creative process of human artists~\cite{hertzmann2022algorithmicPainting}.

\subsection{Painting with High-Level Goals}    
    \subsubsection{Painting from Language Description with Specified Style}

    We use the two loss functions from the StyleCLIPDraw~\cite{schaldenbrand2022styleclipdraw} approach of generating visual content using a style image and language description with our simulated painting environment by concertedly optimizing the Style Objective (Eq. \ref{eq:style} and the Image-Text Similarity Objective (Eq. \ref{eq:img_text}). Results can be seen in Fig. \ref{fig:style_and_text}.
    
    StyleCLIPDraw images tend to lack the original compositional elements of the style image. 
    To retain the style image's composition, we do an initial optimization to replicate the style image. The initial brush stroke plan is now in a local minimum which will be adapted with the full style and text objectives. Fig. \ref{fig:style_and_text} shows that faces and colors appear where they were initially located in the content image thereby providing a method of transferring compositional elements of style. 
    
    \subsubsection{Painting Images Conceptually}
\begin{figure}
        \centering
        \small
        \includegraphics[width=\columnwidth]{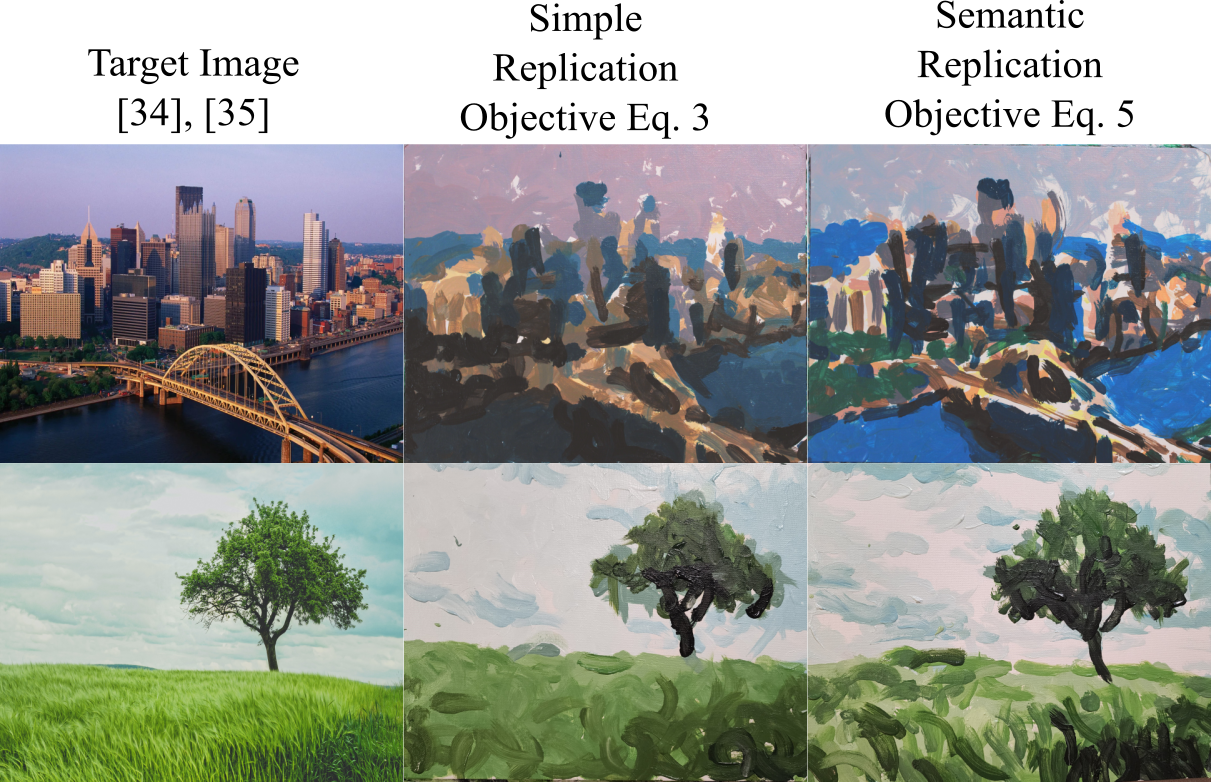}
        \caption{FRIDA's paintings using the Simple Replication Objective (Eq. \ref{eq:l2}) versus the High-Level Semantic Replication Objective (Eq. \ref{eq:clip_conv_loss}).}
        \label{fig:l2_v_clip}\vspace{-7pt}
    \end{figure}

\begin{figure}
        \centering
        \small
        \includegraphics[width=\columnwidth]{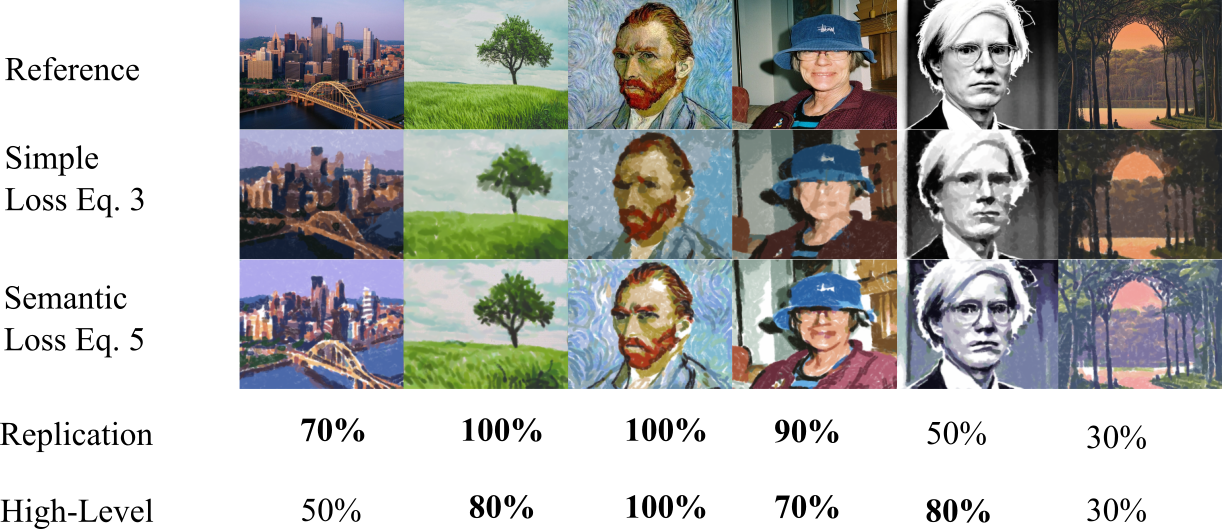}\vspace{-8pt}
        \caption{
        Results from two surveys assessing how well paintings replicated the reference image and how well they retained the high-level content. 
        Percentages shown are preferences for Semantic Loss (Eq.~\ref{eq:clip_conv_loss}) paintings.}
        \label{fig:l2_v_clip_table}\vspace{-15pt}
    \end{figure}

    We compare painting using the Simple and Semantic Replication Objectives in Fig. \ref{fig:l2_v_clip}. 
    We hypothesized that the Semantic Replication Objective would better capture high-level content of the target image. To test this quantitatively, we recruited 103 Amazon Mechanical Turk participants to (1) ``select the painting that looks the most like the target image" and (2) ``select the painting that captures the high-level ideas of the reference image's scene better" and to explain how they made their decision. We refer to these surveys as the replication and high-level questions, respectively. Simulated paintings were used to avoid noise generated by human error in palette preparation of which six pairs were generated with 10 evaluators for each question, painting pair. 73.3\% and 68.3\% of participants selected Semantic Replication Objective paintings for the replication and high-level questions, respectively.  These two averages were both significantly larger than 50\% at a p-value of 0.01 and were not statistically distinct.  While the two questions were different, we noticed that participants claimed to use many of the same features to make their decisions for each question which included colors, shapes, and particular details such as grass and clouds. A breakdown of selections for each painting pair is in Fig.~\ref{fig:l2_v_clip_table}.

    \subsubsection{Sketch to Painting}
    
    User generated sketches can be used to guide the composition of the painting using the Semantic Replication Objective as seen in Fig. \ref{fig:example_results}.
    
    

\section{LIMITATIONS}

    Our system has been simplified in many ways: painting with discrete color options, brush strokes assumed to be independent, no modelling of the wetness of paint, not modeling how much paint is on the brush, and the brush being fixed perpendicular to the canvas.  In future work we hope to explore methods to reduce these limitations.
    
    The second insight of art from \cite{hertzmann2022algorithmicPainting} is that the painting process is adaptable and dynamic enough that new styles can emerge. While our system does adapt its goals as it paints, our definition of dynamic in the system is far more constrained than that of \cite{hertzmann2022algorithmicPainting}. In future work we hope to have a system that can explore more and create novelty as it plans.

\section{CONCLUSIONS}

    We introduce FRIDA, a robotics framework that computationally models the creative process of painting including stochastic nature of acting with paint and a brush. FRIDA is also a robotics initiative to promote human creativity, rather than replacing it, by providing intuitive ways for humans to express their ideas using natural language or sample images. Whereas visual content generation in simulation and that in physical robot painting have been separated in existing work, FRIDA's differential planning environment enables seamless interweaving of various deep neural networks-based content generation models and the actual painting planning. FRIDA's simulation environment developed from real data, an idea known as real2sim2real, reduces the sim2real gap and achieves higher-fidelity to the robot's capabilities than prior learning-enabled painting robots. FRIDA's planning environment enables use of pretrained models as feedback functions, which we have shown can achieve high-level content replication in a survey. FRIDA is open sourced to advocate interdisciplinary research and education in robotics and arts.

\section*{ACKNOWLEDGMENT}

We thank Jia Chen Xu for his contribution to FRIDA's perception, and Jesse Ding and Heera Sekhr for meaningful conversations and domain expertise.
This work was supported by NSF IIS-2112633 and the Technology Innovation Program (20018295, Meta-human: a virtual cooperation platform for a specialized industrial services) funded By the Ministry of Trade, Industry \& Energy (MOTIE, Korea).

\bibliographystyle{IEEEtran}
\bibliography{IEEEabrv,ref}

\appendix 

\subsection{Robot Calibration} \label{sec:calibration}
Minor shifts in the canvas location, different brushes, and altered lighting conditions can greatly divide the simulation from reality.  To account for shifts in the canvas location, a photograph of the canvas is taken, a user uses their mouse to click the corners of the canvas, then the canvas can be isolated from the camera's field of view and oriented properly. There will still be a difference in the coordinate locations of where the robot intends to paint and where it actually paints. To close this gap,
the robot paints 16, small dots evenly distributed across the canvas. The robot then takes a picture of the canvas and computes a homography to translate where the points were intended to be located to where they were actually painted. This homography can be used to ensure that the location that the robot paints on the real canvas corresponds accurately with the simulated canvas.

To calibrate the camera's color and lighting perception, we use a color checker device with 24 colors standard in the photo and video industry. A transformation function is estimated between the perceived color checker values and the true RGB values, and is used for further perception of the canvas.

\begin{figure}
    \centering
    \includegraphics[width=4cm]{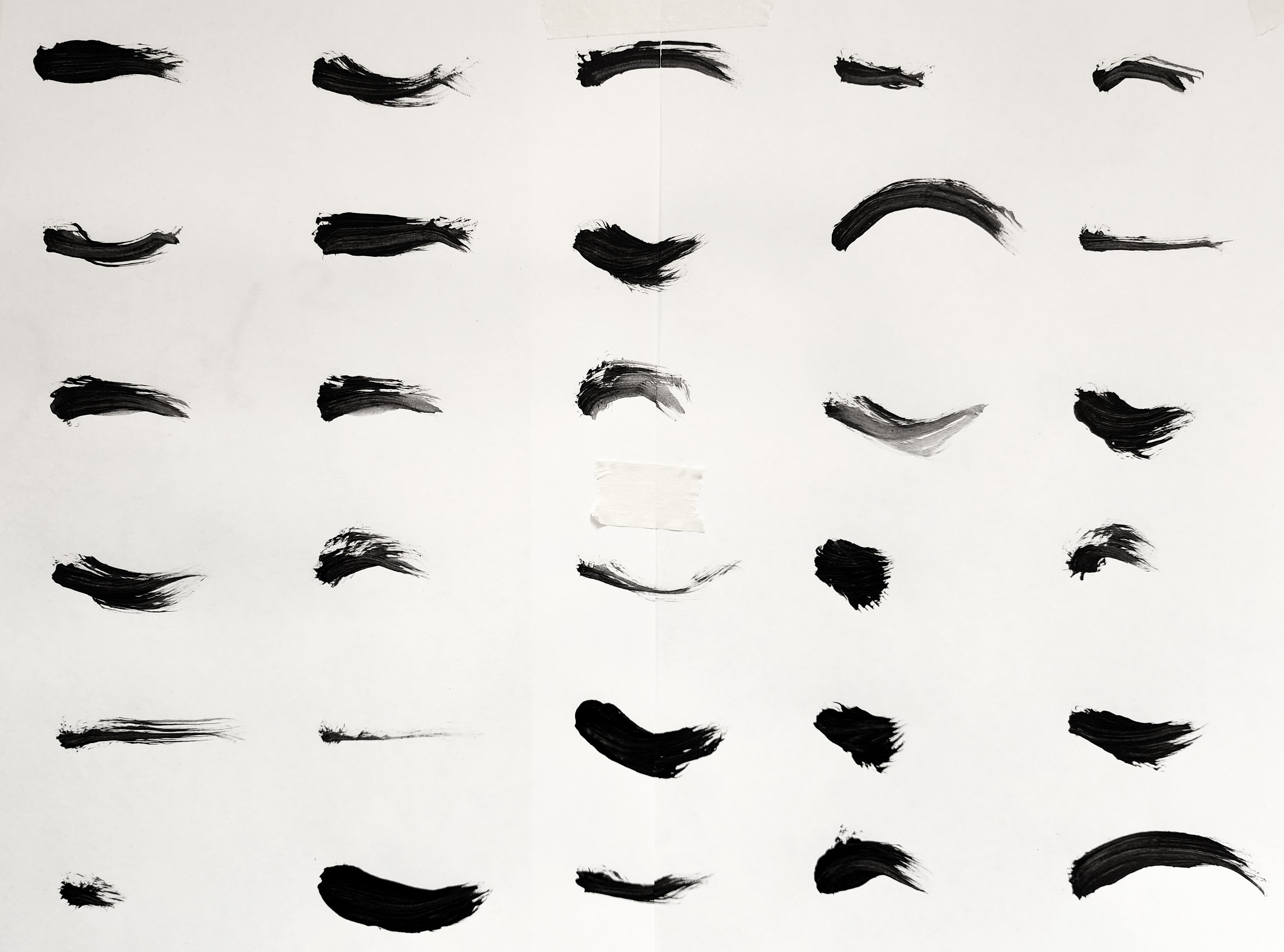}
    \caption{During the calibration phase, the robot makes random brush strokes to create training data for the brush stroke shape model which learns the relationship between the robot parameters and the appearance of a brush stroke.}
    \label{fig:random_brush_strokes}
\end{figure}

When a new brush is attached to the robot, the system needs to determine how far the brush protrudes from the end effector and what the brush is capable of producing when painting. A human operator uses the robot's keyboard interface to lower the brush onto the canvas setting two values: the lightest touch of the brush on the canvas and the hardest brush on the canvas.  Then, the robot makes a series of random brush strokes on the canvas, sampling uniformly from the three brush stroke parameters detailed in Section \ref{sec:brush_shape_model}. Examples of such random brush strokes can be seen in Fig. \ref{fig:random_brush_strokes}.

For brush stroke modeling, the starting and ending points of a stroke have fixed height $h$ values at $0.2$, which corresponds to a light touch of the brush. This is to keep the brush shapes consistent, since when the brush first presses to the page, it is impossible to predict which way the bristles will fan out. Setting a small $h$ ensures that the bristles will not fan out drastically until the stroke moves along its trajectory and it consistently drags behind this trajectory.

\subsection{Brush Stroke Shape Modelling}

\begin{figure}
    \centering
    \includegraphics[width=\columnwidth]{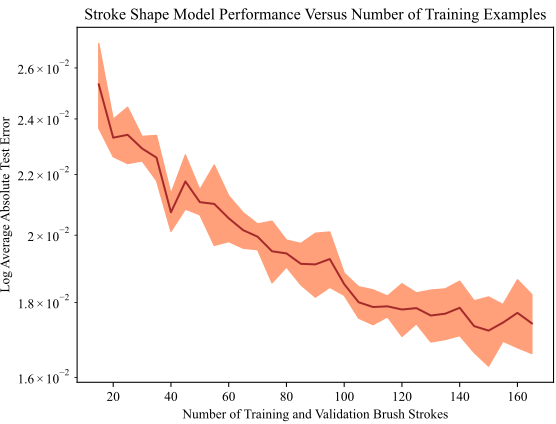}
    \caption{Increasing the number of brush strokes used for training/validation of \texttt{param2stroke} from left-to-right versus the logarithm of the error of the model.}
    \label{fig:param2stroke_error}
\end{figure}

We investigated the number of random strokes needed to accurately train the \texttt{param2stroke} model which translates the stroke shape parameters to the appearance of the stroke. In Fig. \ref{fig:param2stroke_error}, we plot the number of strokes in the training and validation sets versus the logarithm of the mean absolute error on the test set.  The models were tested on a held out set of 42 strokes.  20\% of the stroke-parameter pairs were used for validation to avoid over-fitting \texttt{param2stroke}, and we used five-fold cross validation to compute the log absolute error standard deviation to depict how much the error can fluctuate depending on which random strokes were used for training.  The error generally decreases steadily with more strokes used for training, however, the improvement in performance is small after about 100 training/validation strokes.

We modeled three parameters that affect the shape of the stroke, however, there are many variables that we do not account for because of how challenging it is to perceive them. This noise is depicted in the error standard deviation of Fig.~\ref{fig:param2stroke_error} as well as some qualitative examples of real versus modeled brush strokes in Fig.~\ref{fig:real_v_sim_strokes}. Based on our observations, the most significant unaccounted for variable is the amount of paint on the brush.  As seen in Fig. \ref{fig:random_brush_strokes}, some strokes are lighter or darker depending on how much paint was used. It is a technical challenge to estimate how much paint is on the paint brush and what part of the brush it is on, and we will investigate this in future work. Despite this noise, the model is still capable of learning the most likely appearance of the stroke. In some instances, as depicted in Fig.~\ref{fig:real_v_sim_strokes}, the model's predicted brush stroke looks more likely than the real brush stroke, since the real brush stroke ran out of paint. The model is capable of learning the full, continuous range of stroke shapes accurately, as seen in Fig.~\ref{fig:continuous_strokes}.

\subsection{Paint Colors and Mixing}

FRIDA does not currently support automatic paint mixing.  Instead, FRIDA relies on a human operator to mix a discrete number of paints and provide them.  The paint colors are determined during the initial planning algorithm step (Algorithm~\ref{alg:painting_algorithm}). The brush stroke colors can be clustered with the K-Means algorithm, then displayed to a human operator in a figure similar to that in Fig.~\ref{fig:vs_baseline}. 

Under certain circumstances, color discretization can significantly alter the perceived content of the painting, e.g., a very colorful simulated painting is generated but when discretizing to a small number of colors, the content is no longer apparent.  For this reason, we frequently discretize throughout the optimization process to force the algorithm to operate under the constraint of the limited number of paint colors the user specified.

Painting with an unconstrained order of paint colors leads to two problems: (1) switching colors frequently which creates long execution times spent cleaning the paint brush between colors and (2) touching wet, dark paint on the canvas while painting with a light color and dragging it around to other canvas regions.  To solve both of these problems, we order the brush strokes in the plan from light to dark.  This minimizes the amount of time spent cleaning the brush, and eliminates the ability of the brush incidentally picking up wet, dark paint and spreading it around where light paint should be. Like discretization, ordering the brush strokes by color also affects optimization, so we frequently sort the strokes throughout the optimization process.

\end{document}